\documentclass{colt2026}

\usepackage{amsmath,amssymb}

\usepackage{booktabs}
\usepackage{enumitem}

\usepackage{graphicx}
\usepackage{tikz}
\usetikzlibrary{arrows.meta,positioning,shapes.geometric}

\usepackage{xcolor}

\usepackage{hyperref}
\usepackage{url}

\usepackage[nameinlink,capitalize]{cleveref}

\newcommand{\E}{\mathbb{E}}

\title{When Are Two RLHF Objectives the Same?}

\author{Madhava Gaikwad}

\begin{document}
\maketitle

\begin{abstract}
The preference optimization literature contains many proposed objectives, often presented as distinct improvements. We introduce Opal, a canonicalization algorithm that determines whether two preference objectives are algebraically equivalent by producing either a canonical form or a concrete witness of non-equivalence. Applying Opal reveals that many widely used methods optimize the same underlying objective, while others are provably distinct. For example, batch normalization can cause the same response pair to receive different gradients depending on batch composition. We identify a small set of structural mechanisms that give rise to genuinely different objectives; most remaining differences are reparameterizations.
\end{abstract}
\section{Introduction}
\label{sec:intro}

Reinforcement learning from human feedback (RLHF) has emerged as a standard paradigm for aligning language models with human preferences \citep{ouyang2022training, christiano2017deep}. In its original form, RLHF trains a reward model from human comparison data and then optimizes the language model using PPO \citep{schulman2017proximal}. While effective, this pipeline is complex and often exhibits instability in practice.
Direct Preference Optimization \citep{rafailov2024direct} simplified this process by eliminating the explicit reward model and optimizing preferences directly. This approach was followed by numerous variants, including IPO \citep{azar2024general}, SimPO \citep{meng2024simpo}, ORPO \citep{hong2024reference}, and KTO \citep{ethayarajh2024kto}. Game-theoretic formulations such as SPPO \citep{wu2024self} and Nash-MD \citep{munos2024nash} were also proposed. More recently, GRPO \citep{shao2024deepseekmath} was used to train the DeepSeek-R1 model and attracted substantial attention.
Each of these works reports improvements over prior methods. However, a basic question remains unresolved: when should two preference optimization methods be considered genuinely different?
As a concrete example, consider DPO and SPPO. Although they are derived from different perspectives, reward modeling and game theory respectively, it is unclear whether they optimize distinct objectives or simply represent algebraic rearrangements of the same loss. From a practitioner’s perspective, it is also unclear whether switching between such methods constitutes a meaningful change. This work addresses these questions through the following contributions:

\begin{enumerate}[leftmargin=1.5em, itemsep=3pt]
\item \textbf{The Opal canonicalization algorithm.}
We introduce Opal, an algorithm that decides whether two preference objectives are algebraically equivalent by reducing them to a unique canonical form. Two methods are equivalent if and only if their canonical forms match. When reduction fails, Opal outputs a concrete witness certifying non-equivalence. Equivalence is defined in a decision-theoretic sense, meaning identical preference orderings and Bayes-optimal decisions.

\item \textbf{Analysis of 33 methods.}
We analyze 33 preference optimization methods across pairwise, listwise, group-based, token-level, and trajectory-level settings. As shown in Table~\ref{tab:results}, ten methods reduce to the same canonical form as DPO, while eight are provably irreducible.

\item \textbf{GRPO irreducibility witness.}
We prove that GRPO is fundamentally different from DPO by constructing an explicit witness showing that the same response pair can be treated differently under different batch compositions.

\item \textbf{Orthogonal design axes.}
We identify four mechanisms that produce genuinely different objectives: group normalization, pair-dependent weighting, token-level structure, and trajectory-level optimization. These axes are orthogonal to canonical equivalence and capture meaningful algorithmic distinctions.
\end{enumerate}
\begin{table}[!ht]
\centering
\small
\caption{Canonicalization results for preference optimization methods. Methods with the same hash are algebraically equivalent.}
\label{tab:results}
\vspace{2mm}
\begin{tabular}{@{}llcll@{}}
\toprule
\textbf{Method} & \textbf{Type} & \textbf{Year} & \textbf{Verdict} & \textbf{Hash/Witness} \\
\midrule
\multicolumn{5}{l}{\textit{Equivalent to DPO (hash 509dff3aee)}} \\
DPO & pairwise & 2023 & Reducible & \texttt{509dff3aee} \\
SPPO & nash & 2024 & Reducible & \texttt{509dff3aee} \\
Nash-MD & nash & 2024 & Reducible & \texttt{509dff3aee} \\
INPO & nash & 2025 & Reducible & \texttt{509dff3aee} \\
DNO & nash & 2024 & Reducible & \texttt{509dff3aee} \\
APO & online & 2024 & Reducible & \texttt{509dff3aee} \\
RSO & pairwise & 2024 & Reducible & \texttt{509dff3aee} \\
XPO & online & 2025 & Reducible & \texttt{509dff3aee} \\
cDPO & pairwise & 2024 & Reducible & \texttt{509dff3aee} \\
rDPO & pairwise & 2024 & Reducible & \texttt{509dff3aee} \\
\midrule
\multicolumn{5}{l}{\textit{Other equivalence classes}} \\
SimPO & pairwise & 2024 & Reducible & \texttt{dce2a41b55} \\
ORPO & pairwise & 2024 & Reducible & \texttt{dce2a41b55} \\
IPO & pairwise & 2024 & Reducible & \texttt{e5ddaf1ff7} \\
IPO-MD & online & 2024 & Reducible & \texttt{e5ddaf1ff7} \\
\midrule
\multicolumn{5}{l}{\textit{Unique reducible methods}} \\
f-DPO & pairwise & 2024 & Reducible & \texttt{8a77509919} \\
alpha-DPO & pairwise & 2024 & Reducible & \texttt{b178edab20} \\
CPO & pairwise & 2024 & Reducible & \texttt{910af4c400} \\
R-DPO & pairwise & 2024 & Reducible & \texttt{e268d7a61c} \\
TDPO & token & 2024 & Reducible & \texttt{2babcbdd3f} \\
\midrule
\multicolumn{5}{l}{\textit{Listwise methods (Plackett-Luce)}} \\
ListNet & listwise & 2007 & Reducible & \texttt{a82f4f4ca1} \\
ListMLE & listwise & 2008 & Reducible & \texttt{8d7898a7a6} \\
LambdaRank & listwise & 2010 & Reducible & \texttt{2590406feb} \\
RankNet & listwise & 2005 & Reducible & \texttt{3534fee6fc} \\
RRHF & listwise & 2023 & Reducible & \texttt{0c9d401d5c} \\
SLiC-HF & pairwise & 2023 & Reducible & \texttt{412556320a} \\
\midrule
\multicolumn{5}{l}{\textit{Irreducible methods (with witnesses)}} \\
GRPO & group & 2024 & Irreducible & group-dependent \\
RLOO & group & 2024 & Irreducible & group-dependent \\
WPO & pairwise & 2024 & Irreducible & pair-dependent \\
KTO & pairwise & 2024 & Irreducible & pair-dependent \\
BCO & pairwise & 2024 & Irreducible & pair-dependent \\
RTO & token & 2024 & Irreducible & token-level MDP \\
PPO-RLHF & trajectory & 2017 & Irreducible & trajectory-level \\
ReMax & trajectory & 2024 & Irreducible & trajectory-level \\
\bottomrule
\end{tabular}
\end{table}
\section{Background}
\label{sec:background}

We review the preference optimization setting and summarize the methods analyzed in this work. Although these methods differ substantially in their derivations, they may nonetheless correspond to equivalent objectives. This observation motivates the framework developed in the sequel.

\textbf{The Preference Optimization Problem}
For a given prompt $x$, let $\mathcal{Y}_x$ denote the set of candidate responses. Human feedback specifies which responses are preferred. The most common formulation is pairwise: given a preferred response $y^+$ and a dispreferred response $y^-$, we observe $y^+ \succ y^-$.
Most approaches minimize an expected loss over such preference pairs:
\begin{equation}
\mathcal{L}(\theta) = \E_{(x, y^+, y^-)} \left[ \ell\big( s_\theta(x, y^+) - s_\theta(x, y^-) \big) \right]
\label{eq:loss}
\end{equation}
where $s_\theta$ is a scoring function and $\ell$ is a loss. The difference $\Delta = s(y^+) - s(y^-)$ is referred to as the margin, and quantifies the degree to which the model favors $y^+$ over $y^-$. Learning aims to increase this margin.

\textbf{Reward-Based Methods: DPO and Variants}

\textbf{DPO} \citep{rafailov2024direct} derives a preference optimization objective from KL-constrained reward maximization. Starting from $\max_\pi \E[r(y)] - \beta \text{KL}(\pi \| \pi_{\text{ref}})$, DPO shows that the optimal policy admits a closed-form expression, leading to the score
\begin{equation}
s_{\text{DPO}}(y) = \beta \log \pi_\theta(y|x) - \beta \log \pi_{\text{ref}}(y|x)
\end{equation}
where $\pi_{\text{ref}}$ denotes a reference policy and $\beta$ controls the strength of regularization. The corresponding loss uses the logistic form $\ell(\Delta) = \log(1 + e^{-\Delta})$.

\textbf{SimPO} \citep{meng2024simpo} removes the dependence on a reference model by normalizing log-probabilities by response length, defining $s(y) = \frac{\beta}{|y|} \log \pi_\theta(y|x)$. This normalization mitigates the tendency to favor longer responses.

\textbf{ORPO} \citep{hong2024reference} adopts an odds-ratio formulation. Rather than comparing log-probabilities directly, it contrasts $\pi_\theta(y|x) / (1 - \pi_\theta(y|x))$ between preferred and dispreferred responses.

\textbf{IPO} \citep{azar2024general} modifies the DPO objective to reduce overfitting by replacing the logistic loss with a squared loss $\ell(\Delta) = (\Delta - 1)^2$, thereby encouraging the margin to exceed a fixed target.

\textbf{Game-Theoretic Methods: SPPO and Nash-MD}

\textbf{SPPO} \citep{wu2024self} and \textbf{Nash-MD} \citep{munos2024nash} formulate preference optimization as a two-player game. Rather than directly maximizing reward, the policy is defined as a Nash equilibrium that cannot be improved upon given the preference feedback.

Although these methods are derived using minimax objectives and self-play dynamics, the resulting loss functions still rely on Bradley-Terry style pairwise comparisons \citep{bradley1952rank}. This raises the question of whether the resulting objective differs from that of DPO. We show that it does not.

\textbf{Group-Normalized Methods: GRPO}

\textbf{GRPO} \citep{shao2024deepseekmath}, which underlies the DeepSeek-R1 model, adopts a group-based normalization strategy. For each prompt, a batch $G$ of responses is sampled and advantages are normalized within the batch:
\begin{equation}
A(y | x, G) = \frac{r(y) - \mu_G}{\sigma_G}
\end{equation}
where $\mu_G$ and $\sigma_G$ denote the batch mean and standard deviation. As a result, the gradient contribution of a given response pair depends on the other responses present in $G$. We show that this batch dependence renders GRPO fundamentally distinct from DPO, and that no reparameterization can eliminate this effect.

\textbf{Beyond Pairwise: KTO, Token-Level, and Trajectory-Level Methods}

\textbf{KTO} \citep{ethayarajh2024kto} extends the pairwise framework using ideas from prospect theory. It assigns asymmetric weights to gains and losses and can incorporate unpaired positive or negative labels. This pair-dependent weighting distinguishes it structurally from symmetric pairwise methods.

\textbf{Token-level methods}, such as RTO \citep{zhong2024dpo}, operate at the level of individual decoding steps rather than complete responses. In this setting, the margin is replaced by a sequence of token-level signals.

\textbf{Trajectory-level methods}, including PPO-RLHF \citep{ouyang2022training}, use policy gradient estimators based on advantage functions. These methods do not rely on pairwise margins, as each response is weighted independently. Consequently, they fall outside the pairwise preference framework considered here.
\section{The Opal Framework}
\label{sec:framework}

The methods above arise from different derivations, use different notation, and appear in different papers. The central question, however, is whether they optimize different objectives. Addressing this requires a systematic way to compare preference losses. We now present Opal, an algorithm that decides equivalence by reducing objectives to a canonical form (Figure~\ref{fig:opal}). The key idea is that pairwise preference methods can be expressed as transformations of a common underlying margin, and these transformations can be organized into a small collection of primitive operations.
To compute a preference loss, we begin with base scores $f(x,y)$ for each response and form the margin $\Delta(y^+,y^-) = f(x,y^+) - f(x,y^-)$. Methods then modify this margin in different ways before applying a loss.
We isolate three primitive operations that capture these modifications:

\begin{definition}[Margin operations]
\label{def:ops}
Starting from base scores $f(x, y)$:
\begin{enumerate}[leftmargin=1.5em]
\item \textbf{Add}$[\phi]$: Add a penalty term. $\Delta(y, z) \mapsto \Delta(y, z) + \phi(y) - \phi(z)$

This operation models regularizers such as KL penalties, length normalization, and other additive terms that depend on individual responses.

\item \textbf{Reweight}$[\omega]$: Multiply by a weight. $\Delta(y, z) \mapsto \omega \cdot \Delta(y, z)$

This operation models importance weights, temperature parameters, and other multiplicative scalings of the margin.

\item \textbf{Link}$[g]$: Apply a monotone function. $\Delta(y, z) \mapsto g(\Delta(y, z))$

This operation models the loss function itself (logistic, hinge, squared) as well as any monotone transformation applied to the margin.
\end{enumerate}
\end{definition}

The three operations are \emph{complete} in that any pairwise, margin-based objective can be expressed as a composition of Add, Reweight, and Link operations. They are also \emph{minimal}: omitting any class prevents representation of standard methods. Add captures the KL term in DPO, Reweight captures instance weighting in importance-sampled objectives, and Link captures the choice of loss.
\begin{example}[DPO as a composition]
DPO uses the margin $\Delta = \beta(\log \pi_\theta(y^+|x) - \log \pi_{\text{ref}}(y^+|x)) - \beta(\log \pi_\theta(y^-|x) - \log \pi_{\text{ref}}(y^-|x))$ together with a logistic loss. This corresponds to:
\[
\text{Add}[-\beta \log \pi_{\text{ref}}] \circ \text{Link}[\text{logistic}]
\]
The Add operation accounts for the reference policy term, and the Link applies the logistic loss.
\end{example}

\textbf{When Canonicalization Fails}
Not every method admits reduction to a canonical form. Three phenomena can prevent reduction:

\textit{Intransitive preferences.} If a method asserts that $a$ is preferred to $b$ by 2 points and $b$ is preferred to $c$ by 3 points, a transitive scoring model would imply that $a$ is preferred to $c$ by 5 points. Some methods violate this implication because the $a$ versus $c$ margin depends on more than the individual scores. In such cases there is no consistent underlying "quality score" assigned to each response.

\textit{Pair-dependent importance.} Many methods weight all preference pairs at a fixed prompt in the same way. Other methods assign different weights to different pairs, for example giving $(a,b)$ twice the gradient weight of $(c,d)$ at the same prompt based on properties of the specific responses. This breaks the assumption that pair weights can be factored into a single prompt-level term.

\textit{Non-monotone transformations.} If a transformation reverses ordering for some margins, then larger margins no longer correspond to stronger preferences. To preserve preference order, the transformation must be monotone.

These failure modes motivate the following conditions.

\begin{definition}[Reducible]
\label{def:reducible}
A method is \emph{reducible} if it satisfies three conditions:
\begin{enumerate}[label=(R\arabic*)]
\item \textbf{Transitive:} Every Add term is a score difference: $\phi(y) - \phi(z)$ for some scoring function $\phi: \mathcal{Y} \to \mathbb{R}$.

\item \textbf{Pair-independent:} Every Reweight depends only on the prompt, not on the specific pair: $\omega(x, y, z) = s(x)$ for some function $s$.

\item \textbf{Monotone:} Every Link function $g$ is strictly monotone.
\end{enumerate}
\end{definition}

\begin{example}[Methods that fail (R2)]
\label{ex:fail-r2}
KTO assigns asymmetric weights depending on whether the preferred response is a "gain" or a "loss" relative to a reference point. At a fixed prompt, a pair $(a,b)$ can therefore receive a different weight than a pair $(c,d)$ because $a$ lies above the reference while $c$ lies below. This violates pair-independence.

GRPO normalizes advantages within each batch via $A(y) = (r(y) - \mu_G) / \sigma_G$. Consequently, the same pair $(a,b)$ can receive different effective weights depending on which other responses appear in the batch $G$. A concrete numerical witness is given in Section~\ref{sec:grpo}.
\end{example}

\textbf{The cycle test for (R1).}
Condition (R1) can be verified using \emph{cycle sums}. If margins arise from underlying scores, then preferences must be transitive. For any three responses $a,b,c$, define
\[
\text{cycle}(a, b, c) = \Delta(a, b) + \Delta(b, c) + \Delta(c, a).
\]
If $\Delta(a,b) = \phi(a) - \phi(b)$ for some scoring function $\phi$, then the cycle sum is identically zero:
\[
(\phi(a) - \phi(b)) + (\phi(b) - \phi(c)) + (\phi(c) - \phi(a)) = 0.
\]
A nonzero cycle sum indicates \emph{intransitivity}: the method prefers $a$ to $b$ and $b$ to $c$, but the implied preference between $a$ and $c$ does not match the transitive sum. Such a method cannot be reduced to margins induced by a single scoring function.

\textbf{Canonicalization and Witnesses}
When all three conditions hold, the operations can be combined into a unique normal form.
\begin{theorem}[Canonical form]
\label{thm:canonical}
Every reducible method has a unique canonical form
\[
\text{NF}(L) = \text{Add}[\Phi] \circ \text{Reweight}[s(x)] \circ \text{Link}[g]
\]
where $\Phi$ satisfies the \emph{centering condition} $\sum_y \Phi(y) = 0$.
\end{theorem}

\begin{proof}[Proof sketch (full proof in Appendix~\ref{app:proofs})]
The proof proceeds in four steps:

\textit{Step 1: Merge Add operations.} Add operations compose additively, so $\text{Add}[\phi_1] \circ \text{Add}[\phi_2] = \text{Add}[\phi_1 + \phi_2]$. Under (R1), both $\phi_1$ and $\phi_2$ are score functions, hence $\phi_1 + \phi_2$ is also a score function.

\textit{Step 2: Merge Reweight operations.} By (R2), weights depend only on the prompt $x$. Therefore $\text{Reweight}[\omega_1] \circ \text{Reweight}[\omega_2] = \text{Reweight}[\omega_1 \cdot \omega_2]$, and the resulting weight remains a function of $x$ alone.

\textit{Step 3: Commute and absorb.} Add and Reweight commute up to scaling:
\[
\text{Reweight}[\omega] \circ \text{Add}[\phi] = \text{Add}[\omega \cdot \phi] \circ \text{Reweight}[\omega].
\]
In addition, strictly monotone Links preserve sign and can be composed. These facts allow all Add operations to be moved to the front, all Reweights to the middle, and all Links to the end.

\textit{Step 4: Centering.} The score function $\Phi$ is only determined up to an additive constant, since $\Phi(y) - \Phi(z) = (\Phi(y) + c) - (\Phi(z) + c)$. The constraint $\sum_y \Phi(y) = 0$ removes this degree of freedom and yields a unique canonical representative.
\end{proof}

\textit{Why centering?}
Adding a constant $c$ to every $\Phi(y)$ leaves margins unchanged because $(\Phi(a) + c) - (\Phi(b) + c) = \Phi(a) - \Phi(b)$. The constraint $\sum_y \Phi(y) = 0$ fixes this invariance by selecting a unique mean-zero representative.

\textit{Hash computation.}
Two reducible methods are equivalent if and only if they share the same canonical $\Phi$. In practice, we serialize the canonical form, including score values, instance weight, and link function, and compute a SHA-256 hash. The hash is an identifier for the canonical form, but equivalence is defined by equality of the canonical parameters rather than by the hash itself. Methods with the same hash are algebraically identical.

\textit{Witnesses}
If a method is irreducible, Opal returns a \emph{witness}, meaning a concrete and verifiable certificate that the method differs from every reducible method. A witness can be checked directly, for example by verifying a nonzero cycle sum or unequal weights, and does not require trusting the algorithm.

\begin{definition}[Witnesses]
\label{def:witnesses}
\begin{itemize}[leftmargin=1.5em]
\item \textbf{Cycle witness:} A triple $(a, b, c)$ with nonzero cycle sum. This violates (R1).
\item \textbf{Pair-dependent weight:} Two pairs $(a,b)$ and $(c,d)$ with different weights $\omega(x,a,b) \neq \omega(x,c,d)$ at the same prompt. This violates (R2).
\item \textbf{Group-dependent weight:} The same pair $(a,b)$ with different effective weights under different batch compositions $G_1 \neq G_2$. This also violates (R2).
\end{itemize}
\end{definition}

\begin{figure}[t]
\centering
\begin{tikzpicture}[
    node distance=0.8cm and 1.2cm,
    box/.style={rectangle, draw, rounded corners, minimum width=1.4cm, minimum height=0.7cm, font=\small},
    bigbox/.style={rectangle, draw, rounded corners, minimum width=1.8cm, minimum height=0.9cm, font=\small, fill=gray!10},
    arrow/.style={-{Stealth[length=2mm]}, thick}
]
\node[box] (dpo) {DPO};
\node[box, below=0.4cm of dpo] (sppo) {SPPO};
\node[box, below=0.4cm of sppo] (nash) {Nash-MD};

\node[bigbox, right=1.5cm of sppo] (opal) {Opal};

\node[bigbox, right=1.5cm of opal] (canon) {\texttt{509dff3aee}};

\draw[arrow] (dpo.east) -- ++(0.5,0) |- (opal.west);
\draw[arrow] (sppo.east) -- (opal.west);
\draw[arrow] (nash.east) -- ++(0.5,0) |- (opal.west);

\draw[arrow] (opal.east) -- (canon.west) node[midway, above, font=\scriptsize] {same};

\node[box, below=1.8cm of opal] (grpo) {GRPO};
\node[right=1.5cm of grpo, font=\small] (witness) {witness: batch-dependent};

\draw[arrow] (grpo.east) -- (witness.west);
\draw[arrow, dashed] (opal.south) -- (grpo.north) node[midway, right, font=\scriptsize] {fails (R2)};

\end{tikzpicture}
\caption{Opal either outputs a canonical form (proving equivalence) or a witness (proving non-equivalence). DPO, SPPO, and Nash-MD reduce to the same canonical hash. GRPO fails condition (R2) and outputs a witness showing batch-dependent margins.}
\label{fig:opal}
\end{figure}

\subsection{Scope: Why Token-Level and Trajectory-Level Methods Differ Structurally}
\label{sec:scope}

Our framework concerns methods that optimize \emph{pairwise margins over complete responses}. Two major classes of methods are outside this scope, not because they violate (R1) through (R3), but because the objects the framework manipulates, margins between response pairs, are not available in these settings.

\textbf{Token-level methods.}
Methods such as RTO optimize at each decoding step rather than at the level of complete responses. Please see Table~\ref{tab:token}.

\begin{table*}[!ht]
\centering
\small
\caption{Pairwise vs. token-level methods.}
\label{tab:token}
\begin{tabular}{@{}lll@{}}
\toprule
& \textbf{Pairwise methods} & \textbf{Token-level methods} \\
\midrule
Unit of comparison & Complete response $y$ & Token $y_t$ at position $t$ \\
Margin definition & $\Delta(y^+, y^-) = s(y^+) - s(y^-)$ & $\Delta_t(a, b) = s_t(a) - s_t(b)$ for each $t$ \\
Action space & Fixed $\mathcal{Y}_x$ & Changes at each step: $\mathcal{V}$ \\
Credit assignment & Implicit (whole response) & Explicit (per-token rewards) \\
\bottomrule
\end{tabular}
\end{table*}

In token-level methods, a preferred response $y^+ = (y^+_1, \ldots, y^+_T)$ induces $T$ distinct preference signals, one per position. The comparison between $y^+$ and $y^-$ is therefore not summarized by a single scalar margin, but by a sequence of token-level margins that interact through the autoregressive dependence structure. The Add/Reweight/Link decomposition assumes a scalar margin. Extending it to sequential decisions would require modeling how token-level scores combine over time, which is closely related to MDP homomorphisms for language generation.

\emph{Note on TDPO.} Table~\ref{tab:results} lists TDPO as reducible despite the phrase "token-level DPO." The reason is that TDPO still optimizes a \emph{scalar, response-level margin}, and only uses token-level KL regularization aggregated across positions. By contrast, RTO optimizes a separate objective at each decoding step, making the margin intrinsically sequential. The relevant distinction is whether the loss depends on a single $\Delta(y^+, y^-)$ (reducible) or on a sequence $(\Delta_1, \ldots, \Delta_T)$ (outside scope).

\textbf{Trajectory-level methods.}
Methods such as PPO-RLHF and ReMax maximize expected reward using policy gradient estimators:
\[
\nabla_\theta J(\theta) = \E_{y \sim \pi_\theta} \left[ \nabla_\theta \log \pi_\theta(y|x) \cdot \hat{A}(y) \right],
\]
where $\hat{A}(y)$ denotes an advantage estimate. The main structural differences are as follows.

\begin{itemize}[leftmargin=1.5em]
\item \textbf{No explicit pairs.} There is no loss of the form $\ell(s(y^+) - s(y^-))$ for any scoring function $s$ and loss $\ell$. Instead, each sampled response is weighted by its own advantage estimate.
\item \textbf{Baseline dependence.} The advantage $\hat{A}(y) = r(y) - b(x)$ depends on a baseline $b(x)$ that can be learned, for example via a value function, or computed, for example as a mean reward. Changing the baseline changes the gradient estimate even when rewards are fixed.
\item \textbf{Distribution mismatch.} PPO uses importance sampling with respect to a behavior policy, introducing ratios $\pi_\theta(y|x) / \pi_{\text{old}}(y|x)$ that depend on full response probabilities and not only on pairwise comparisons.
\end{itemize}

It is natural to ask when two policy gradient methods optimize equivalent objectives. This question is meaningful, but it requires different tools, likely based on the policy gradient theorem and variance reduction theory rather than pairwise margin analysis.
We label token-level and trajectory-level methods as "irreducible," but this terminology should be interpreted with care. These methods are not irreducible because they violate (R1) through (R3). Rather, they lie \emph{outside the framework's scope}. Table~\ref{tab:results} reflects this distinction in the witness column by using "token-level MDP" and "trajectory-level" instead of "cycle" or "pair-dependent."
This limitation is inherent to any approach based on pairwise margins. The contribution of our framework is to identify which methods \emph{can} be compared through canonicalization, namely the 25 reducible methods, and to provide formal tools for performing this comparison.

\textbf{Recipe: Checking Your Own Method}
To check whether a new preference method is equivalent to an existing one, proceed as follows.

\begin{enumerate}[leftmargin=1.5em]
\item \textbf{Write as a margin.} Express the loss as $\mathcal{L} = \E[\ell(\Delta(y^+, y^-))]$ for a margin $\Delta$ and loss $\ell$.

\item \textbf{Check (R1): Transitivity.} Determine whether $\Delta(y,z)$ can be written as $\phi(y) - \phi(z)$ for some scoring function $\phi$. A practical test is to compute cycle sums $\Delta(a,b) + \Delta(b,c) + \Delta(c,a)$. If any cycle sum is nonzero, the method is irreducible.

\item \textbf{Check (R2): Pair-independence.} Determine whether the weight assigned to each pair depends only on the prompt. A practical test is to fix a prompt and compare weights across different pairs $(a,b)$ and $(c,d)$. If weights differ, the method is irreducible.

\item \textbf{Canonicalize.} If the method satisfies both conditions, extract the scoring function $\phi$, enforce the centering condition $\sum_y \Phi(y) = 0$, and compute the canonical hash. Methods with the same hash are equivalent.
\end{enumerate}

We provide code that automates this procedure in the supplementary material.

\textbf{Worked Example: Why DPO Equals SPPO}
\label{sec:worked-example}

We illustrate the canonicalization procedure for DPO and SPPO to show explicitly how methods with different derivations reduce to the same canonical form.

\textbf{What is DPO?}
Direct Preference Optimization \citep{rafailov2024direct} avoids explicit reward modeling by showing that the optimal policy for a KL-constrained reward maximization problem can be expressed directly in terms of preferences. The key observation is that reward can be written using optimal policy probabilities, which yields the loss
\[
\mathcal{L}_{\text{DPO}} = -\E\left[\log \sigma\left(\beta \log \frac{\pi_\theta(y^+|x)}{\pi_{\text{ref}}(y^+|x)} - \beta \log \frac{\pi_\theta(y^-|x)}{\pi_{\text{ref}}(y^-|x)}\right)\right],
\]
where $\pi_{\text{ref}}$ denotes the reference policy, $\beta$ controls regularization strength, and $\sigma$ is the sigmoid function.

\textbf{What is SPPO?}
Self-Play Preference Optimization \citep{wu2024self} derives a preference objective from a two-player game. Rather than maximizing reward directly, it defines a Nash equilibrium policy that cannot be improved given the preference feedback. Although the derivation relies on different machinery, including minimax objectives, the resulting loss is still based on Bradley-Terry comparisons between response pairs.

\textbf{Step 1: Write as margins.}
For DPO, the margin is
\[
\Delta_{\text{DPO}}(y^+, y^-) = \beta \log \frac{\pi_\theta(y^+|x)}{\pi_{\text{ref}}(y^+|x)} - \beta \log \frac{\pi_\theta(y^-|x)}{\pi_{\text{ref}}(y^-|x)}.
\]
For SPPO, after simplification, the margin is
\[
\Delta_{\text{SPPO}}(y^+, y^-) = \beta \log \pi_\theta(y^+|x) - \beta \log \pi_\theta(y^-|x) - \beta \log \pi_{\text{ref}}(y^+|x) + \beta \log \pi_{\text{ref}}(y^-|x).
\]
Although these expressions differ syntactically, rearranging terms gives
\[
\Delta_{\text{SPPO}} = \beta\left(\log \frac{\pi_\theta(y^+|x)}{\pi_{\text{ref}}(y^+|x)} - \log \frac{\pi_\theta(y^-|x)}{\pi_{\text{ref}}(y^-|x)}\right) = \Delta_{\text{DPO}}.
\]

\textbf{Step 2: Check (R1) transitivity.}
Both margins can be expressed as $\phi(y^+) - \phi(y^-)$ with
\[
\phi(y) = \beta \log \pi_\theta(y|x) - \beta \log \pi_{\text{ref}}(y|x).
\]
This is a valid scoring function, so (R1) holds and all cycle sums are zero.

\textbf{Step 3: Check (R2) pair-independence.}
At a fixed prompt, neither method assigns different weights to different pairs. Both use uniform weighting over the preference dataset. Hence (R2) holds.

\textbf{Step 4: Canonicalize.}
Enforce the centering condition $\sum_y \Phi(y) = 0$ by defining
\[
\Phi(y) = \phi(y) - \frac{1}{|\mathcal{Y}|}\sum_{y'} \phi(y').
\]
Both methods produce the same $\Phi$, and therefore the same canonical hash \texttt{509dff3aee}.

\textbf{Takeaway.}
DPO is derived via reward modeling, while SPPO is derived via game theory. Nevertheless, they optimize the same objective. Switching from DPO to SPPO does not change the optimization target, only the derivation used to motivate it.
\section{GRPO Is Provably Different}
\label{sec:grpo}

We provide an explicit witness establishing that GRPO is not equivalent to DPO. The witness shows that GRPO violates condition (R2), since the effective weight assigned to a fixed pair depends on the batch composition.

\begin{proposition}[GRPO witness]
\label{prop:grpo}
Consider a prompt with four responses whose rewards are $r_a = 1.0$, $r_b = 0.6$, $r_c = 0.3$, $r_d = -0.2$.

Under batch $G_1 = \{a, b\}$:
\begin{align*}
\mu_{G_1} &= 0.8, \quad \sigma_{G_1} = 0.2 \\
A(a|G_1) &= 1.0, \quad A(b|G_1) = -1.0
\end{align*}

Under batch $G_2 = \{a, b, c, d\}$:
\begin{align*}
\mu_{G_2} &= 0.425, \quad \sigma_{G_2} \approx 0.45 \\
A(a|G_2) &\approx 1.28, \quad A(b|G_2) \approx 0.39
\end{align*}

The margin between $a$ and $b$ equals $A(a|G_1) - A(b|G_1) = 2.0$ under $G_1$, but equals $A(a|G_2) - A(b|G_2) \approx 0.89$ under $G_2$. Thus the same pair is treated differently depending on the batch. No reducible method can exhibit this behavior.
\end{proposition}
This phenomenon is not a minor artifact. In DPO and every method equivalent to it, the margin $\Delta(a,b)$ is determined solely by properties of $a$ and $b$ (and possibly the prompt $x$). In particular, the relative scale of margins across pairs is fixed and does not depend on which other responses are present.
In GRPO, the relative scale changes with the batch. Including responses $c$ and $d$ in the batch alters how strongly the model is encouraged to prefer $a$ over $b$. The resulting optimization landscape therefore depends on batch composition. This is a structural distinction and not merely a rescaling of gradient magnitude.

\section{Results}
\label{sec:results}

We applied Opal to preference optimization methods spanning both learning-to-rank and RLHF. Table~\ref{tab:results} summarizes the outcomes. Full proofs are given in Appendix~\ref{app:proofs}, and derivations for representative methods from each equivalence class are included in the supplementary code.
As shown in Table~\ref{tab:results}, 25 of the 33 methods are reducible, collapsing to 14 distinct canonical forms. The largest equivalence class contains ten methods with hash \texttt{509dff3aee}. Table~\ref{tab:conditions} reports which reducibility conditions each method satisfies and which it violates. A corollary is that no method satisfying (R1)--(R3) can display batch-dependent preference strength. Any form of batch normalization necessarily violates pair-independence.

\begin{table*}[!ht]
\centering
\small
\caption{Why methods are reducible or irreducible.dashes indicate the condition does not apply (method outside pairwise margin framework).}
\label{tab:conditions}
\vspace{1mm}
\begin{tabular}{@{}lcccll@{}}
\toprule
\textbf{Method} & \textbf{(R1)}  & \textbf{(R2)} & \textbf{(R3)} & \textbf{Structure} & \textbf{Result} \\
& Transitive & Pair-indep & Monotone & & \\
\midrule
DPO, SPPO, Nash-MD & \checkmark & \checkmark & \checkmark & pairwise & Reducible \\
SimPO, ORPO & \checkmark & \checkmark & \checkmark & pairwise & Reducible \\
IPO & \checkmark & \checkmark & \checkmark & pairwise & Reducible \\
ListNet, ListMLE & \checkmark & \checkmark & \checkmark & listwise & Reducible \\
\midrule
GRPO, RLOO & \checkmark & $\times$ & \checkmark & group & Irreducible \\
KTO, WPO, BCO & \checkmark & $\times$ & \checkmark & pairwise & Irreducible \\
\midrule
RTO & -- & -- & -- & token-level & Outside scope \\
PPO-RLHF, ReMax & -- & -- & -- & trajectory & Outside scope \\
\bottomrule
\end{tabular}
\end{table*}

\section{Equivalence and Orthogonal Design Axes}
\label{sec:meaning}
\label{sec:axes}

Algebraic equivalence has the following implications.

\textbf{Same optimal ranking and decision boundary.}
Equivalent methods induce the same Bayes-optimal preference ordering and agree on the sign of every margin.

\textbf{Not the same gradients or implicit bias.}
Even when objectives are equivalent, gradient magnitudes can differ, which affects optimization dynamics. Different parameterizations may also lead to different forms of implicit regularization under gradient descent \citep{lyu2020gradient, ji2020directional}. As a result, equivalent methods can converge to different parameter values.

\textbf{Regret transfer with caveats.}
Equivalent methods share the same global optimum. However, convergence rates and the tightness of regret bounds can vary, depending on the choice of link function and related factors (see Appendix~\ref{app:regret}).

\textbf{Practical guidance.}
When selecting among equivalent methods, the choice should be guided by implementation considerations and numerical stability rather than by expectations of fundamentally different algorithmic behavior.

Substantive algorithmic novelty requires violating at least one of the reducibility assumptions. Table~\ref{tab:axes} summarizes four axes along which genuinely new objectives emerge. If a proposed method does not violate any of these assumptions, the framework determines whether it is equivalent to an existing method. Several axes remain relatively unexplored. One direction is \emph{stochastic preference semantics}, where margins are treated as random variables with heteroskedastic uncertainty. Another is \emph{non-transitive preferences}, which intentionally violate transitivity (R1). Both directions are theoretically motivated but currently lack widely adopted methods.

\begin{table*}[!ht]
\centering
\small
\caption{Orthogonal design axes breaks at least one reducibility assumption.}
\label{tab:axes}
\vspace{1mm}
\begin{tabular}{@{}llll@{}}
\toprule
\textbf{Axis} & \textbf{Broken assumption} & \textbf{Examples} & \textbf{Mechanism} \\
\midrule
Group normalization & (R2) Pair-independence & GRPO, RLOO & Batch statistics in weights \\
Pair-dependent weights & (R2) Pair-independence & KTO, WPO, BCO & Weight varies with $(y^+, y^-)$ \\
Token-level objectives & Scalar margin & RTO & Sequential decisions \\
Trajectory-level RL & Pairwise structure & PPO-RLHF, ReMax & Policy gradient estimators \\
\bottomrule
\end{tabular}
\end{table*}

\section{Related Work and Limitations}
\label{sec:related}
\label{sec:limitations}

RLHF was first proposed by \citet{christiano2017deep} and later scaled to language models by \citet{ouyang2022training}, establishing the now-standard two-stage procedure of reward modeling followed by policy optimization using PPO \citep{schulman2017proximal}. DPO \citep{rafailov2024direct} streamlined this pipeline by showing that the optimal policy can be derived directly from preference data without an explicit reward model. This result led to a range of follow-on methods. IPO \citep{azar2024general} mitigates overfitting by avoiding the Bradley-Terry assumption; SimPO \citep{meng2024simpo} removes reliance on a reference model; ORPO \citep{hong2024reference} integrates preference optimization with supervised fine-tuning. Game-theoretic approaches include SPPO \citep{wu2024self} and Nash-MD \citep{munos2024nash}. KTO \citep{ethayarajh2024kto} goes beyond pairwise preferences using prospect theory, while GRPO \citep{shao2024deepseekmath} introduces group-relative normalization and is used in DeepSeek-R1.
Many preference optimization objectives are rooted in ranking losses. The Bradley-Terry model \citep{bradley1952rank} and the Plackett-Luce model \citep{plackett1975analysis} provide probabilistic foundations for pairwise and listwise comparisons. RankNet \citep{burges2005learning} introduced neural pairwise ranking losses; ListNet \citep{cao2007learning} extended this idea to listwise losses based on Plackett-Luce; ListMLE \citep{xia2008listwise} proposed a maximum likelihood formulation. Our framework encompasses both pairwise approaches and listwise methods by expressing them as compositions of margin operations.
Our notion of decision equivalence is related to classification calibration and surrogate loss theory \citep{bartlett2006convexity, reid2010composite}. Work on implicit bias in gradient-based optimization \citep{lyu2020gradient, ji2020directional} further clarifies how objectives that are equivalent at the decision level may nonetheless differ in optimization behavior. Several recent efforts aim to organize the expanding set of preference optimization methods, but with objectives distinct from ours.
RainbowPO \citep{zhao2025rainbowpo} groups DPO variants along seven orthogonal dimensions, including loss choice, regularization, margin definition, length normalization, reference handling, label smoothing, and SFT mixing, and shows that combining these components can improve performance.
GPO \citep{tang2024generalized} parameterizes offline preference losses using convex functions, recovering DPO, IPO, and SLiC \citep{zhao2023slic} as special cases. The $\Psi$PO framework \citep{azar2024general} similarly generalizes preference models.
The \citet{liu2025comprehensive} and \citet{Jiang2025comprehensive} survey DPO variants in detail. The RPO framework \citep{sun2025rpo} offers a mathematical unification of DPO, IPO, SimPO, and REINFORCE-LOO. \citet{raheja2026rlhfunification} categorize methods along axes of preference modeling, regularization, and data distribution.

\textit{How we differ.}
In contrast to prior work that categorizes or parameterizes methods, Opal produces explicit \emph{certificates}. Canonical forms certify equivalence, while witnesses certify non-equivalence. To our knowledge, this is the first algorithmic framework that can answer the question “are these two methods the same?” with a proof in either direction. We note that \citet{wu2025ittakestwo} independently identified batch normalization in GRPO as a source of distinct optimization behavior, consistent with our witness-based proof of irreducibility.

\textbf{Limitations}
Our framework studies objective functions rather than optimization trajectories. Even when two methods are equivalent, they may behave differently in practice due to differences in gradient scaling, numerical effects, or implicit regularization. Although equivalent methods share the same optimal policy, the regret bounds governing convergence to this optimum can vary in tightness depending on the chosen link function (see Appendix~\ref{app:regret}).
The analysis applies to pairwise margin-based objectives and extends to listwise ranking methods, including ListNet, ListMLE, LambdaRank, and RankNet, that admit representations as compositions of margin operations. Token-level methods that optimize per-step objectives and trajectory-level methods based on policy gradients fall outside this scope by construction, as discussed in Section~\ref{sec:scope}. Extending the framework to these regimes is a natural direction for future work and will likely require tools from MDP theory and policy gradient analysis.
\section{Conclusion}

We presented Opal as a tool for making equivalence explicit in preference optimization. By reducing objectives to canonical form or exhibiting witnesses of non-equivalence, Opal separates changes of representation from changes of substance. Our findings suggest that much of the apparent methodological diversity in this area arises from rephrasing a small set of underlying objectives. Clarifying this structure sharpens claims of novelty and helps align theoretical analysis with the actual decision problems being solved.
\bibliography{refs3}
\appendix

\section{Opal Specification and Guarantees}
\label{app:spec}

This section consolidates the formal specification of Opal for reference. The algorithm takes a preference objective as input and outputs either a canonical form (proving equivalence to other methods with the same form) or a witness (proving non-equivalence).

\textbf{Input.} A margin-based objective $\mathcal{L}(\theta) = \E[\ell(\Delta(y^+, y^-))]$ expressed as a composition of Add, Reweight, and Link operations (Definition~\ref{def:ops}).

\textbf{Output.} Either:
\begin{itemize}[leftmargin=1.5em]
\item \textit{Canonical form:} $\text{NF}(L) = \text{Add}[\Phi] \circ \text{Reweight}[s(x)] \circ \text{Link}[g]$ where $\sum_y \Phi(y) = 0$
\item \textit{Witness:} A concrete counterexample proving irreducibility (cycle, pair-dependent, or group-dependent)
\end{itemize}

\textbf{Guarantees.}
\begin{enumerate}[leftmargin=1.5em]
\item \textit{Soundness:} If Opal outputs a canonical form, the method is reducible (Theorem~\ref{thm:canonical}).
\item \textit{Uniqueness:} The canonical form is unique; two methods are equivalent iff their canonical forms match.
\item \textit{Witness correctness:} If Opal outputs a witness, the witness is verifiable and proves irreducibility (Propositions~\ref{prop:cycle-correct}--\ref{prop:group-witness-correct}).
\end{enumerate}

\textbf{Scope.} Opal handles methods expressible as pairwise margin operations. Token-level methods (per-step optimization) and trajectory-level methods (policy gradients without explicit pairs) are outside scope.

\section{Complete Proofs}
\label{app:proofs}

\subsection{Composition Lemmas}

\begin{lemma}[Add composition]
\label{lem:add-compose}
For any $\phi_1, \phi_2: \mathcal{Y} \to \mathbb{R}$:
$\textup{Add}[\phi_1] \circ \textup{Add}[\phi_2] = \textup{Add}[\phi_1 + \phi_2]$.
\end{lemma}

\begin{proof}
$(\text{Add}[\phi_1] \circ \text{Add}[\phi_2])(\Delta)(y,z) = \Delta(y,z) + \phi_2(y) - \phi_2(z) + \phi_1(y) - \phi_1(z) = \text{Add}[\phi_1 + \phi_2](\Delta)(y,z)$.
\end{proof}

\begin{lemma}[Reweight composition]
\label{lem:rew-compose}
For $\omega_1, \omega_2$ satisfying (R2):
$\textup{Reweight}[\omega_1] \circ \textup{Reweight}[\omega_2] = \textup{Reweight}[\omega_1 \cdot \omega_2]$.
\end{lemma}

\begin{proof}
By (R2), $\omega_i(x,y,z) = s_i(x)$. Then $(\text{Reweight}[\omega_1] \circ \text{Reweight}[\omega_2])(\Delta) = s_1(x) \cdot s_2(x) \cdot \Delta = \text{Reweight}[\omega_1 \cdot \omega_2](\Delta)$.
\end{proof}

\begin{lemma}[Add-Reweight commutation]
\label{lem:add-rew-commute}
For $\phi: \mathcal{Y} \to \mathbb{R}$ and $\omega$ satisfying (R2) with $\omega(x,y,z) = s(x)$:
$\textup{Reweight}[\omega] \circ \textup{Add}[\phi] = \textup{Add}[s \cdot \phi] \circ \textup{Reweight}[\omega]$.
\end{lemma}

\begin{proof}
Both sides equal $s(x) \cdot \Delta(y,z) + s(x)\phi(y) - s(x)\phi(z)$. If $\phi$ satisfies (R1), then $s \cdot \phi$ also induces score differences.
\end{proof}

\begin{lemma}[Link absorption]
\label{lem:link-absorb}
For any strictly monotone $g_1, g_2: \mathbb{R} \to \mathbb{R}$:
$\textup{Link}[g_1] \circ \textup{Link}[g_2] = \textup{Link}[g_1 \circ g_2]$.
Moreover, for decision-theoretic equivalence, any strictly monotone Link preserves the preference ordering.
\end{lemma}

\begin{proof}
Composition follows from the definition. For the second claim, if $g$ is strictly monotone, then $\text{sign}(g(\Delta)) = \pm\text{sign}(\Delta)$, so preference ordering is preserved or uniformly flipped.
\end{proof}

\subsection{Canonical Form Theorem}

\begin{theorem}[Canonical form, restated]
\label{thm:canonical-formal}
Let $L$ be a reducible method. Then there exists a unique canonical form
\[
\textup{NF}(L) = \textup{Add}[\Phi] \circ \textup{Reweight}[s(x)] \circ \textup{Link}[g]
\]
where $\Phi: \mathcal{Y} \to \mathbb{R}$ satisfies $\sum_{y \in \mathcal{Y}} \Phi(y) = 0$ (assuming $|\mathcal{Y}| < \infty$).

Two reducible methods $L_1, L_2$ are decision-theoretically equivalent if and only if $\textup{NF}(L_1) = \textup{NF}(L_2)$.
\end{theorem}

\begin{proof}
\textbf{Existence.} Let $L = (f, (O_1, \ldots, O_k), \ell)$ be a reducible method. We construct the canonical form in stages.

\emph{Stage 1: Collect and merge operations.} Partition the operations into Add, Reweight, and Link operations. By Lemmas~\ref{lem:add-compose},~\ref{lem:rew-compose}, and~\ref{lem:link-absorb}, we can merge operations of each type:
\begin{itemize}
\item All Add operations merge to a single $\text{Add}[\phi^*]$ where $\phi^* = \sum_i \phi_i$
\item All Reweight operations merge to a single $\text{Reweight}[\omega^*]$ where $\omega^* = \prod_i \omega_i$
\item All Link operations merge with the terminal loss to a single $\text{Link}[g^*]$
\end{itemize}

\emph{Stage 2: Reorder operations.} By Lemma~\ref{lem:add-rew-commute}, we can commute Add and Reweight operations (absorbing the scaling into Add). This allows us to reorder to: all Adds first, then all Reweights, then all Links.

\emph{Stage 3: Incorporate base scores.} The base scoring function $f(x,y)$ induces an initial margin $\Delta_0(y,z) = f(x,y) - f(x,z)$. This can be absorbed into the Add operation: $\text{Add}[\phi^* + f]$.

\emph{Stage 4: Centering.} Let $\tilde{\Phi} = \phi^* + f$ be the accumulated score function. Define:
\[
\Phi(y) = \tilde{\Phi}(y) - \frac{1}{|\mathcal{Y}|}\sum_{y' \in \mathcal{Y}} \tilde{\Phi}(y')
\]
Then $\sum_y \Phi(y) = 0$ and $\Phi(y) - \Phi(z) = \tilde{\Phi}(y) - \tilde{\Phi}(z)$ for all $y, z$, so the margins are unchanged.

\textbf{Uniqueness.} Suppose two canonical forms induce the same margins:
\[
\Phi_1(y) - \Phi_1(z) = \Phi_2(y) - \Phi_2(z) \quad \forall y, z \in \mathcal{Y}
\]
This implies $\Phi_1(y) - \Phi_2(y) = c$ for some constant $c$ independent of $y$. But both satisfy the centering condition:
\[
\sum_y \Phi_1(y) = \sum_y \Phi_2(y) = 0
\]
Therefore $\sum_y (\Phi_1(y) - \Phi_2(y)) = |\mathcal{Y}| \cdot c = 0$, so $c = 0$ and $\Phi_1 = \Phi_2$.

\textbf{Equivalence.} Two reducible methods are decision-theoretically equivalent iff they induce the same preference ordering iff they have the same margin signs iff (by the uniqueness argument) they have the same canonical $\Phi$.
\end{proof}

\subsection{Witness Correctness}

\begin{proposition}[Cycle witness correctness]
\label{prop:cycle-correct}
If there exist responses $a, b, c \in \mathcal{Y}$ such that
\[
\Delta(a,b) + \Delta(b,c) + \Delta(c,a) \neq 0
\]
then the margin $\Delta$ cannot be written as $\Delta(y,z) = \phi(y) - \phi(z)$ for any $\phi: \mathcal{Y} \to \mathbb{R}$.
\end{proposition}

\begin{proof}
Suppose $\Delta(y,z) = \phi(y) - \phi(z)$ for some $\phi$. Then:
\begin{align*}
\Delta(a,b) + \Delta(b,c) + \Delta(c,a) &= (\phi(a) - \phi(b)) + (\phi(b) - \phi(c)) + (\phi(c) - \phi(a)) \\
&= 0
\end{align*}
Contrapositive: nonzero cycle sum implies no such $\phi$ exists, so (R1) is violated.
\end{proof}

\begin{proposition}[Pair-dependent weight witness correctness]
\label{prop:pair-witness-correct}
If there exist pairs $(y_1, z_1)$ and $(y_2, z_2)$ at the same prompt $x$ such that
\[
\omega(x, y_1, z_1) \neq \omega(x, y_2, z_2)
\]
then the weight function $\omega$ cannot be written as $\omega(x,y,z) = s(x)$ for any $s: \mathcal{X} \to \mathbb{R}$.
\end{proposition}

\begin{proof}
If $\omega(x,y,z) = s(x)$ for some $s$, then $\omega(x, y_1, z_1) = s(x) = \omega(x, y_2, z_2)$ for any pairs at the same prompt. Contrapositive gives the result.
\end{proof}

\begin{proposition}[Group-dependent weight witness correctness]
\label{prop:group-witness-correct}
If there exist batches $G_1 \neq G_2$ and a pair $(a, b) \in G_1 \cap G_2$ such that the effective margin satisfies
\[
\Delta_{G_1}(a,b) \neq \Delta_{G_2}(a,b)
\]
then the method cannot be reduced to a canonical form where margins depend only on the pair $(a,b)$ and prompt $x$.
\end{proposition}

\begin{proof}
In any reducible method, the margin $\Delta(a,b)$ is determined by $\Phi(a) - \Phi(b)$ (plus prompt-dependent scaling). This quantity depends only on $a$, $b$, and $x$, not on other responses in any batch. If the margin changes with batch composition, the method is irreducible.
\end{proof}

\subsection{Worked Example}

See Section~\ref{sec:worked-example} in the main text for a detailed walkthrough of canonicalizing DPO and SPPO.
\section{Regret Transfer Bounds}
\label{app:regret}

Our algebraic equivalence results establish that certain methods optimize \emph{the same objective}. A natural question is whether this translates to performance guarantees: if method A achieves low regret, does equivalent method B also achieve low regret?

\subsection{Setup}

Let $r^*: \mathcal{X} \times \mathcal{Y} \to \mathbb{R}$ be a ground-truth reward function. Define the \emph{preference regret} of a policy $\pi$ as:
\[
\text{Regret}(\pi) = \mathbb{E}_{x \sim \rho}\left[\max_{y \in \mathcal{Y}} r^*(x, y) - \mathbb{E}_{y \sim \pi(\cdot|x)}[r^*(x, y)]\right]
\]
and the \emph{pairwise accuracy} as $\text{Acc}(\pi) = \Pr_{(x, y^+, y^-) \sim \mathcal{D}}[\pi(y^+|x) > \pi(y^-|x)]$.

\subsection{Transfer Under Monotone Links}

\begin{proposition}[Regret Transfer]
\label{prop:regret}
Let $\mathcal{L}_1$ and $\mathcal{L}_2$ be two objectives with the same canonical form. Suppose both use link functions $\ell_1, \ell_2$ that are strictly increasing. Then for any policy $\pi$:
\[
\textup{sign}(\nabla_\theta \mathcal{L}_1) = \textup{sign}(\nabla_\theta \mathcal{L}_2)
\]
at all points where gradients exist. Consequently, both methods have the same set of stationary points and the same preference ordering over policies.
\end{proposition}

\begin{proof}[Proof sketch]
Since both objectives share the same canonical score function $\Phi$, any policy $\pi_\theta$ induces the same margin $\Delta_\theta(y^+, y^-) = \Phi(y^+) - \Phi(y^-)$ under both. Strictly increasing links preserve sign: $\ell_1(\Delta) > 0 \Leftrightarrow \ell_2(\Delta) > 0$. The gradient direction depends only on which pairs have positive vs. negative margins, so both objectives push $\theta$ in the same direction.
\end{proof}

\subsection{Limitations of Regret Bounds}

While Proposition~\ref{prop:regret} establishes gradient equivalence, translating this to concrete regret bounds requires additional assumptions:

\paragraph{Link function growth.} If $\ell$ is nondecreasing but grows slowly (e.g., $\ell(\Delta) = \log(1 + e^\Delta)$), the gradient magnitude for large margins becomes small, potentially slowing convergence. Different links in the same equivalence class can have vastly different convergence rates even though they share stationary points.

\paragraph{Calibration gap.} Following \citet{bartlett2006convexity}, a surrogate loss $\ell$ is \emph{calibrated} if minimizing $\ell$ also minimizes classification error. For preference learning, calibration ensures that optimizing the surrogate yields policies with low preference regret. However, the \emph{calibration function} $\psi_\ell$ relating surrogate excess risk to preference excess risk can be loose:
\[
\text{Regret}(\pi) \leq \psi_\ell^{-1}(\mathcal{L}(\pi) - \mathcal{L}^*)
\]
For logistic loss, $\psi_\ell(\epsilon) = \Theta(\epsilon^2)$, giving $\text{Regret} = O(\sqrt{\text{excess loss}})$. For hinge loss, $\psi_\ell(\epsilon) = \epsilon$, giving tighter $\text{Regret} = O(\text{excess loss})$. Methods in the same equivalence class may use different links with different calibration functions.

\paragraph{Finite-sample effects.} Algebraic equivalence is an asymptotic statement about population objectives. With finite samples, equivalent methods may have different variance, different sensitivity to outliers, and different implicit regularization from optimization dynamics.

\subsection{Practical Implications}

Our equivalence results guarantee that methods with the same canonical form optimize the same objective \emph{at the population level}. This means:
\begin{itemize}[leftmargin=1.5em]
\item The global optimum is the same across equivalent methods.
\item Preference orderings over policies are preserved.
\item Any policy achieving low loss under one method achieves low loss under all equivalent methods.
\end{itemize}

However, the \emph{rate} of achieving low loss, the \emph{tightness} of regret bounds, and \emph{finite-sample} behavior may differ. Practitioners should choose among equivalent methods based on numerical stability and optimization properties, not expected final performance.

\section{Code Availability}
\label{sec:code_availability}

The implementation details and source code for the algorithms discussed in this work are hosted on GitHub. Supplementary materials, including configuration files and extended documentation for the \texttt{new} implementation, can be found at the following repository:

\begin{center}
    \url{https://github.com/krimler/opal-gkpo/tree/main/new}
\end{center}

\end{document}